\documentclass[a4paper,conference]{IEEEtran}
\usepackage{cite}
\usepackage{amsmath,amssymb,amsfonts}
\usepackage{algorithmic}
\usepackage[dvipdfmx]{graphicx}
\usepackage{textcomp}
\usepackage{xcolor}
\usepackage{bm}
\usepackage{comment}
\usepackage{here}

\def\BibTeX{{\rm B\kern-.05em{\sc i\kern-.02IEEEtran5em b}\kern-.08em
    T\kern-.1667em\lower.7ex\hbox{E}\kern-.125emX}}
\begin{document}

\title{Improving Minimal Gated Unit for Sequential Data}

\author{
\IEEEauthorblockN{1\textsuperscript{st} Kazuki Takamura}
\IEEEauthorblockA{\textit{Kanazawa University}\\
Kakuma-machi 920-1192, Kanazawa, Japan \\
ktakamura@csl.ec.t.kanazawa-u.ac.jp}\\
\and
\IEEEauthorblockN{2\textsuperscript{nd} Satoshi Yamane}
\IEEEauthorblockA{\textit{Kanazawa University}\\
Kakuma-machi 920-1192, Kanazawa, Japan \\
syamane@is.t.kanazawa-u.ac.jp}
}

\maketitle

\begin{abstract}
In order to obtain a model which can process sequential data related to machine translation and speech recognition faster and more accurately, we propose adopting Chrono Initializer as the initialization method of Minimal Gated Unit. We evaluated the method with two tasks: adding task and copy task. As a result of the experiment, the effectiveness of the proposed method was confirmed.
\end{abstract}

% \begin{IEEEkeywords}
% Minimal Gated Unit, Chrono Initializer.
% \end{IEEEkeywords}

%研究背景的な何か。なんでこの研究が必要なの？
\section{INTRODUCTION}
Recently, Long Short-Term Memory (LSTM)\cite{lstm} has been successful in various fields. For example, machine translation and speech recognition are well known.
Moreover, there are researches to find a simpler and more compact model than LSTM such as Gated Recurrent Unit (GRU)\cite{gru} and Minimal Gated Unit (MGU)\cite{mgu}.
The advantages of these models are the clarity of structure, the reduction of the number of parameters, and the shortening of learning time.
On the other hand, they have a problem that the accuracy of the model becomes worse because the number of parameters are reduced.\par
In this paper, in order to obtain a model which can process sequential data faster and more accurately, we change the initialization method of MGU which has a relatively small number of parameters.

% 近年, Long Short-Term Memory(LSTM)が様々な分野で成果を上げ、LSTMに関する研究が数多く存在する。
% その研究の一つとして、Gated Recurrent Unit(GRU), Minimal Gated Unit(MGU) のようなLSTMに比べてよりシンプルでコンパクトなモデルを求める研究がある。
%これらのモデルの利点は、構造の明快さ、パラメータ量の削減、および学習時間の短縮などである。
% 一方、パラメータ数が少なくなるため、モデルの精度が悪くなるという問題がある\par
% 本研究ではより精度の高いモデルを求めるため,比較的パラメータの数が少ないMGUの初期化方法を変更する.

\section{RELATED WORKS}
\subsection{Minimal Gated Unit}
Minimal Gated Unit\cite{mgu} is a derivative of GRU\cite{gru}. In detail, Reset gate and Input gate in GRU are integrated to one gate.

%Minimal Gated Unit (MGU)\cite{b2}はGRUの派生モデルであり,GRUのResetゲートとUpdateゲートを統合した単一のゲートでモデルに流れる情報を管理する.\par

\begin{figure}[htbp]
    \centerline{\includegraphics[width=0.7\hsize]{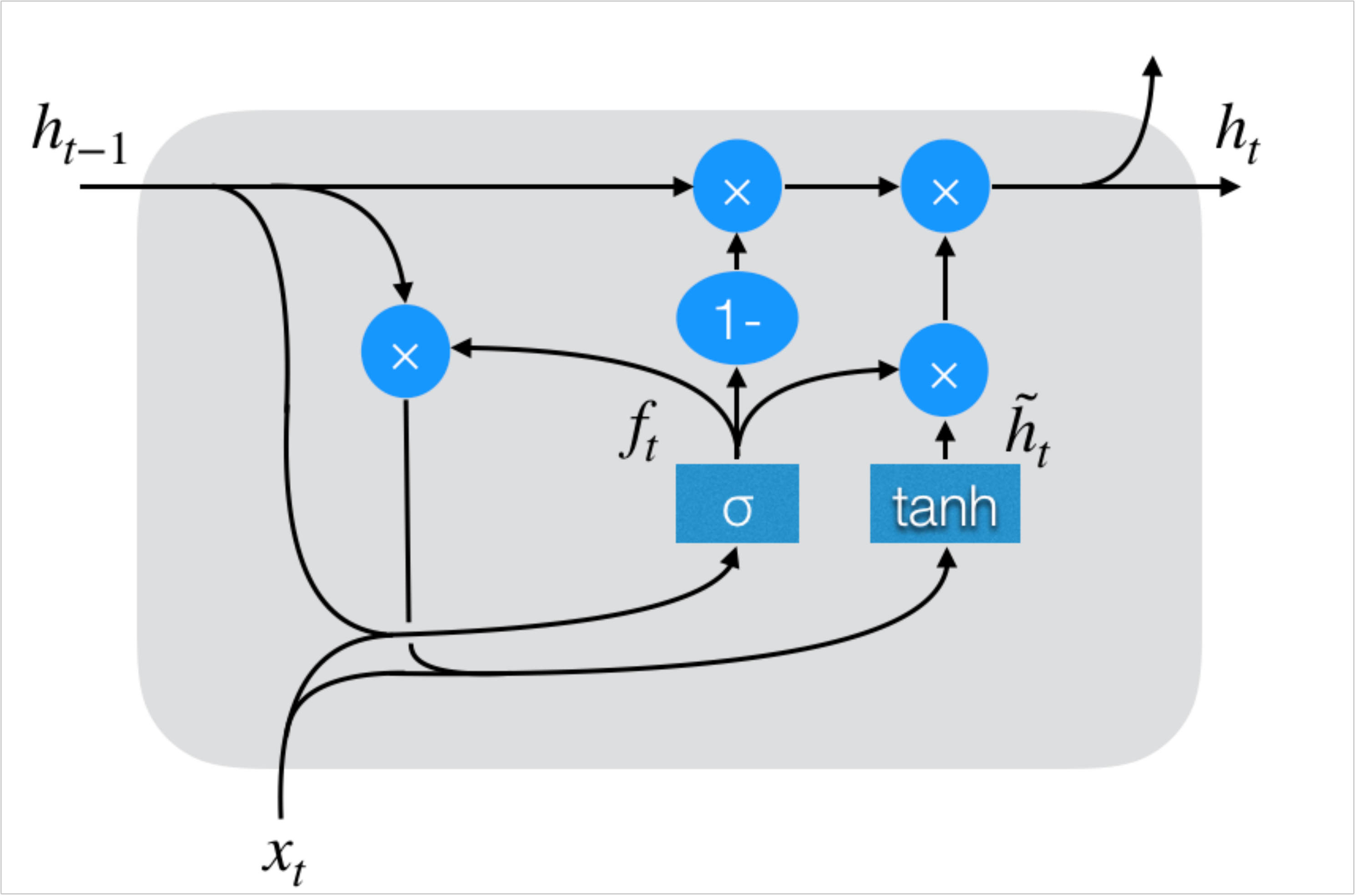}}
    \caption{Minimal Gated Unit}
    \label{MGU}
\end{figure}

MGU is expressed by the following equations, where the weight for hidden state is $\left(\bm{W}^{(f)}_h,\bm{W}_h\right)$, the weight for inputs is $\left(\bm{W}^{(f)}_x,\bm{W}_x\right)$, the bias is $\left(\bm{b}^{(f)},\bm{b}\right)$.

%隠れ状態に対する重みを(W(f)h;Wh),入力に対する重みを(W(f)x;Wx),バイアスを(b(f);b)とした時に,MGUは以下の式で表される.

\begin{eqnarray}
    \bm{f}_t &=& \sigma{\left(\bm{W}^{(f)}_h\bm{h}_{t-1}+\bm{W}^{(f)}_x\bm{x}_t+\bm{b}^{(f)}\right)} \label{eq:gate_mgu}\\
    \tilde{\bm{h}_t} &=& \tanh{\left(\bm{W}_h(\bm{f}_t\odot\bm{h}_{t-1})+\bm{W}_x\bm{x}_t+\bm{b}\right)} \label{eq:tilde_hidden_mgu}\\
    \bm{h}_t &=& (1-\bm{f}_t)\odot\bm{h}_{t-1}+\bm{f}_t\odot\tilde{\bm{h}_t} \label{eq:hidden_mgu}
\end{eqnarray}

\subsection{Chrono Initializer}
Chrono Initializer\cite{chrono} was proposed as a gate bias initialization method for LSTM.
This initializer makes LSTM better capture long-term dependence of data.\par
Chrono Initializer is expressed by the following equations, where $U$ is the uniform distribution and $T_{max}$ is the expected range of long-term dependencies to be captured.

% Chrono Initializer~\cite{b3}は,LSTMにデータの長期依存性をよりうまく捉えさせるためのゲートのバイアスの初期化方法として提案された.\par
% when $\bm{U}$ is the uniform distribution and $\bm{T_{max}}$ is the expected range of long-term dependencies to be captured, Chrono Initializerは式(\ref{eq:chrono_forget}),(\ref{eq:chrono_input})のように表される.

\begin{eqnarray}
    \bm{b}^{(f)}&\sim&\log(U([1,T_{max}-1])) \label{eq:chrono_forget}\\
    \bm{b}^{(i)}&=&-\bm{b}^{(f)} \label{eq:chrono_input}
\end{eqnarray}

\section{PROPOSED METHOD AND COMPARISON}
We propose adopting Chrono Initializer\cite{chrono} as the initialization method of MGU\cite{mgu}.
By using Chrono Initializer, it is expected that long-term dependency can be easily grasped by MGU and the performance will be improved.
The equations of our proposed method are expressed by adding Eq. (\ref{eq:chrono_forget}) to Eq. (\ref{eq:gate_mgu}), (\ref{eq:tilde_hidden_mgu}), and (\ref{eq:hidden_mgu}).
We compare the proposed method with the method whose gate bias value is set to 1.
For convenience, we call this method MGU (Const.).

% MGUの初期化方法としてChrono Initializerを採用することを提案する。
% Chrono Initializerを採用することによって、より長期依存性捉えやすくなり、性能が上がると考えられる。
% 式でいうと,提案手法は、MGUの式\ref{eq:gate_mgu},\ref{eq:tilde_hidden_mgu},\ref{eq:hidden_mgu}にChrono Initializerの式\ref{eq:chrono_forget}を加える形になる。
% 今回はゲートのバイアスを1設定した場合(便宜上,従来手法と呼ぶ)との比較を行う。

\section{EXPERIMENTS AND RESULTS}

\subsection{Methods}

We evaluate the performance of the model with two tasks: adding task and copy task\cite{task}.\par
Adding task is a task that adds real numbers.
The input consists of two rows.
The first row is a random real-value column, and the second row is a mask of 0 and 1 (1 appears only twice).
The output is the value obtained by adding the real-value in the column where the mask value is 1.
In this task, we set the length of column to 50 and 250.\par

Copy task is a task to copy a sequence.
When T is given, the length of input and output is T+20 and these components are integers from 0 to 9.
The input consists of 10 random values (from 1 to 8), T-1 dummies (0), 1 signal (9), and 10 dummies in order.
The output consists of T+10 dummies and first 10 random values of the input in order.
In this task, we set T to 50 and 200.

%今回はadding taskとcopy taskの2つのタスクで提案手法の性能を測る。
%adding taskは実数の加算を行うタスクである. 入力は 2 行の列からなり,1 行目はランダムな実数列,2 行目は0と1のマスクビット(ただし 1 は二つだけ)で構成される. 
% 出力はマスクビットが 1の列の実数の値を加算して得られる値となる.このタスクでは列の長さを50,250に設定した。
% 平均二乗誤差を評価方法とする.

% copy taskは長さ10の数列のコピーを行うタスク。Tが与えられた時,入力と出力はともにT+20の長さからなる。
% 入力は長さ10のランダムな数列,長さT-1のダミー(0),信号となる一つの数字(9),長さ10のダミーの順番で構成される.
% 出力はT+10,10で構成される。
% このタスクではTを50,200に設定した。
% ソフトマックスクロスエントロピーを評価方法とする。

\subsection{Results}
{
\setlength\floatsep{0pt}
\setlength\abovecaptionskip{0pt}
\setlength\intextsep{0pt}
\setlength\textfloatsep{0pt}
\begin{figure}[H]
    \centerline{\includegraphics[clip,width=0.85\hsize]
    {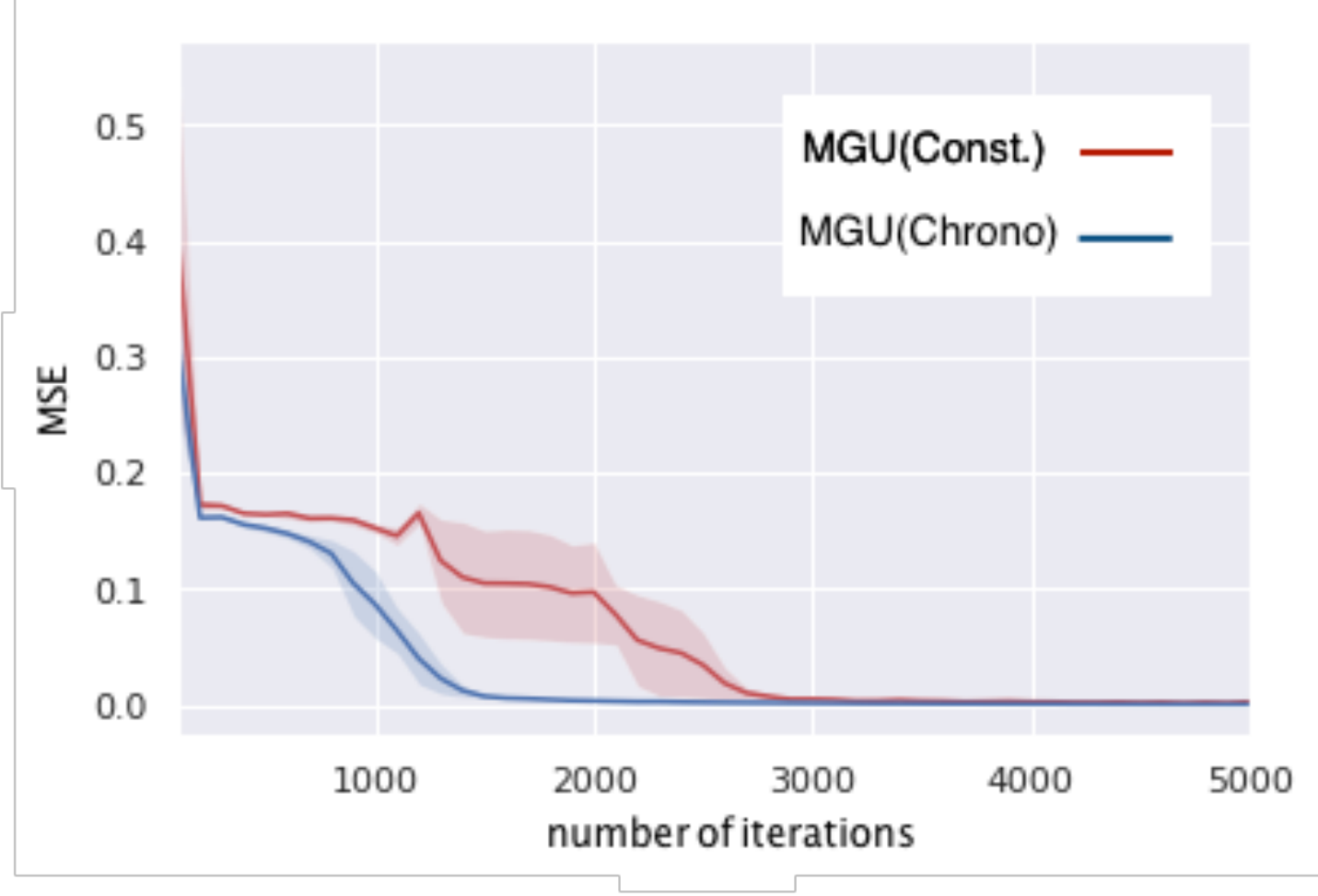}}
    \caption{Adding task result (length=50)}
    \label{adding 50}
\end{figure}
\begin{figure}[H]
    \centerline{\includegraphics[clip,width=0.85\hsize]{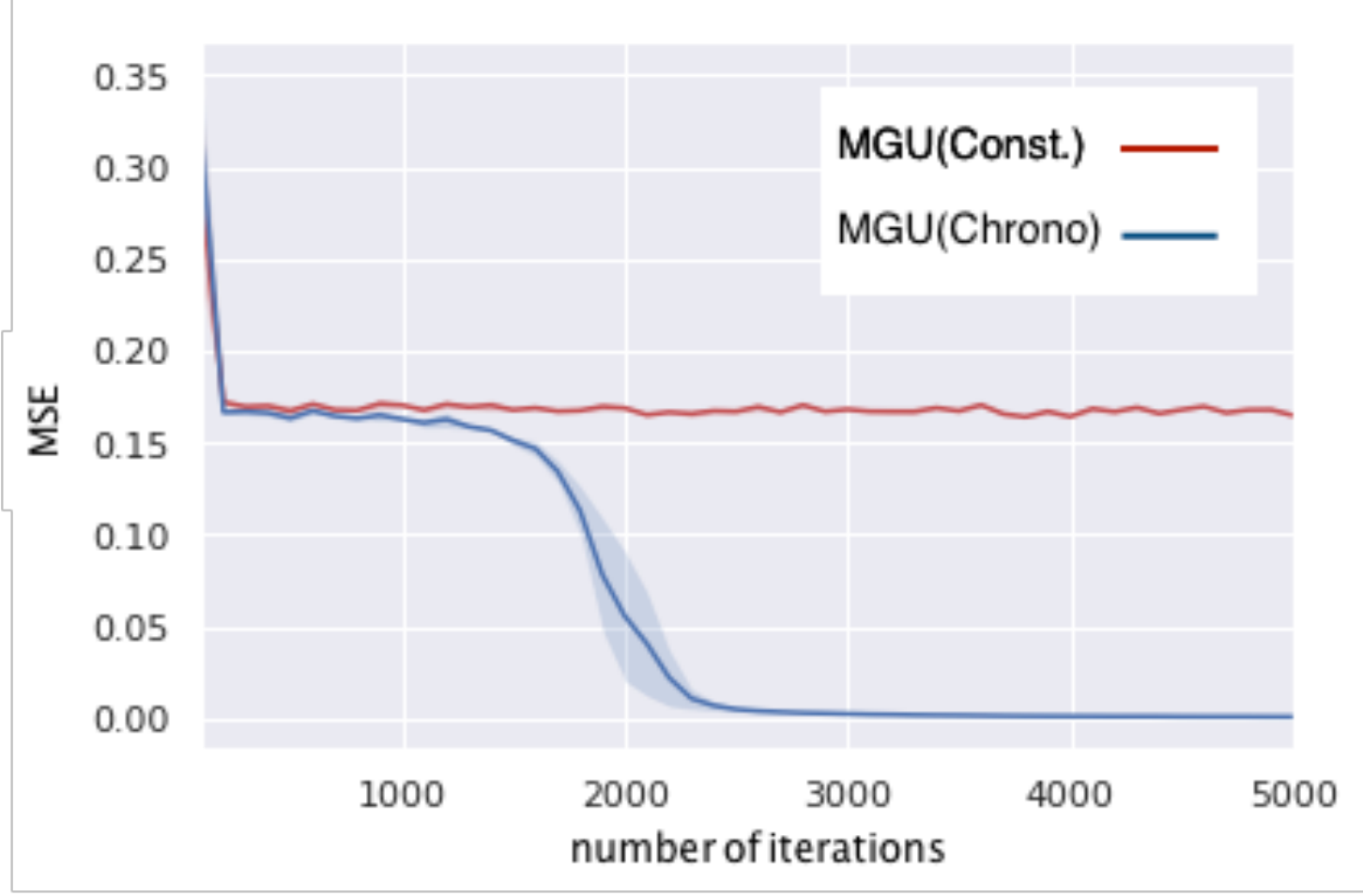}}
    \caption{Adding task result (length=250)}
    \label{adding 250}
\end{figure}
\begin{figure}[H]
    \centerline{\includegraphics[clip,width=0.85\hsize]{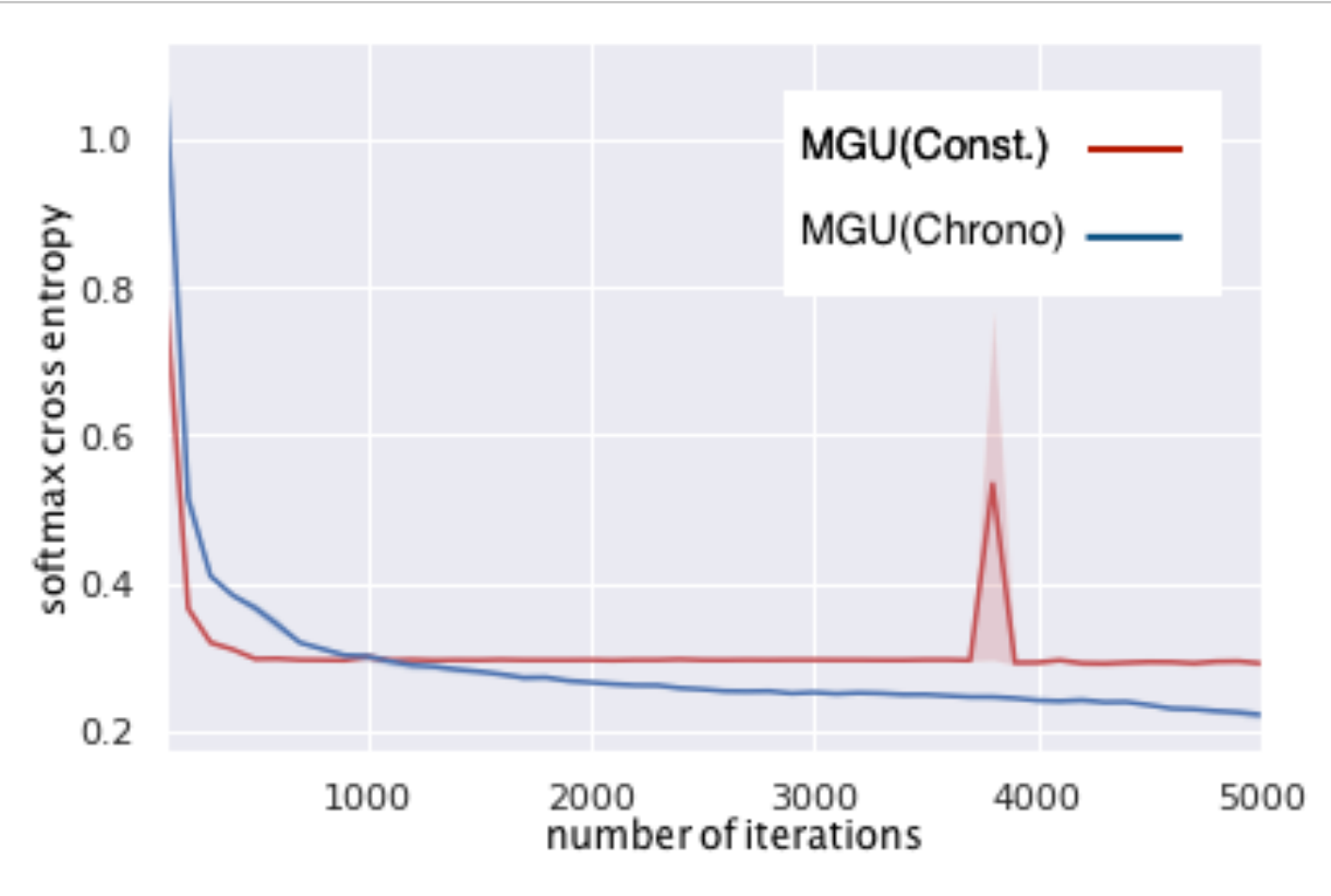}}
    \caption{Copy task result (T=50)}
    \label{copy 50}
\end{figure}
\begin{figure}[H]
    \centerline{\includegraphics[clip,width=0.85\hsize]{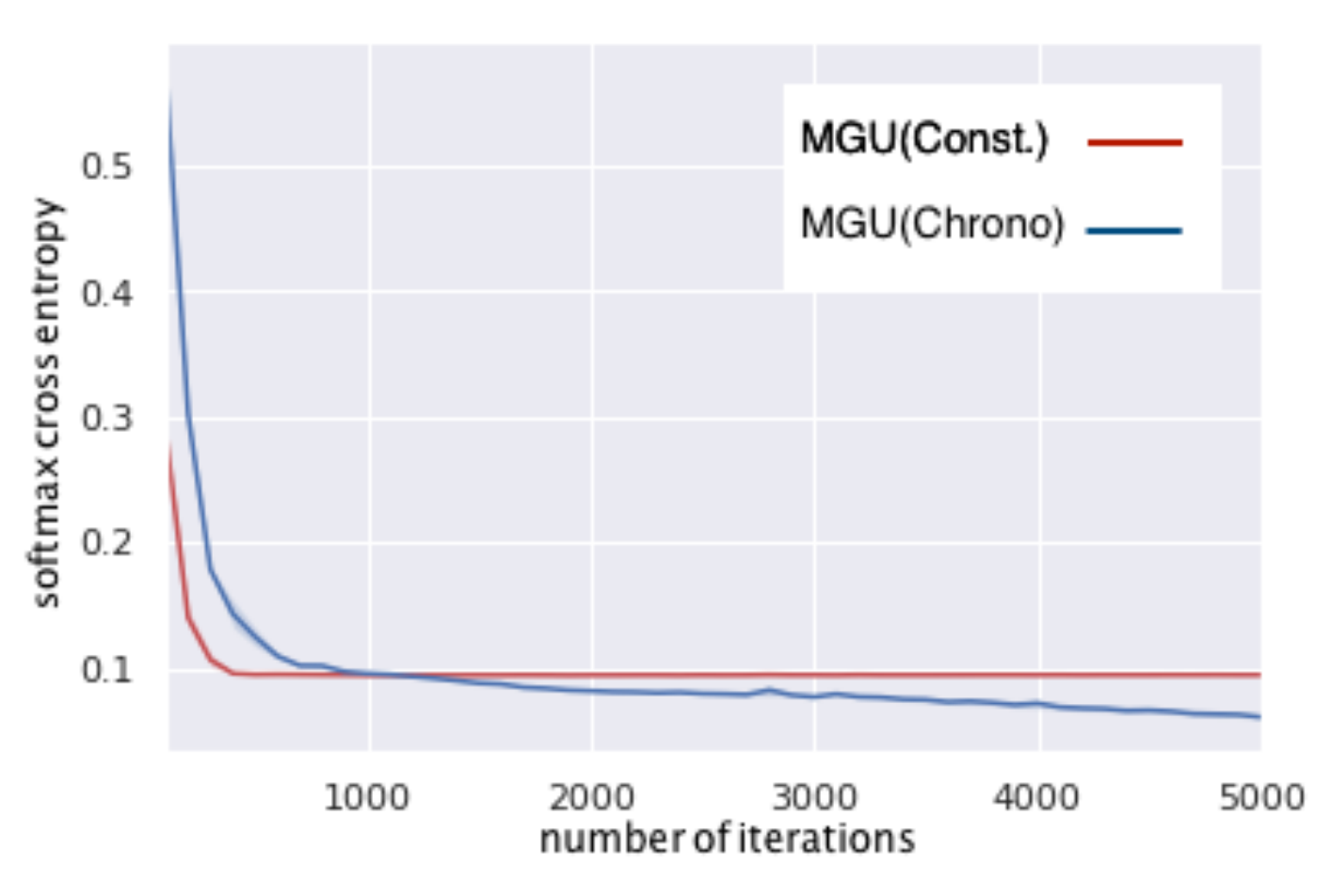}}
    \caption{Copy task result (T=200)}
    \label{copy 200}
\end{figure}
}

\begin{table}[H]
    \caption{Parameters for experiment}
    \begin{center}
      \begin{tabular}{|c|c|c|c|c|}
        \hline
        & \textbf{iteration}& \textbf{batch size}& \textbf{hidden size}& \textbf{learning rate}\\ \hline
        \textbf{adding}& 5,000& 50& 128& 0.001\\ \hline
        \textbf{copy}& 5,000& 128& 128& 0.001\\ \hline
      \end{tabular}
      \label{parameter}
    \end{center}
\end{table}

Table \ref{parameter} represents the parameters of our experiment.
We experimented three times for each previous method and our proposed method. 
The solid lines represent the average of the values, and the colored ranges represent the degree of variation.\par

%adding task, copy taskともにiteration*batch size分のデータをランダムに生成した。
% 提案手法と従来手法のそれぞれで同じ実験を3回ずつ行った.
% それぞれのグラフの実線は3回の実験で得られた値の平均を表し,色のついた範囲はデータのばらつきの度合いを表す.また,赤い線が従来手法の結果,青い線が提案手法の結果を表す.

Fig. \ref{adding 50} and \ref{adding 250} are the results of the adding task.
The vertical axis represents Mean Squared Error (MSE), and the horizontal axis represents the number of iterations.
From Fig. \ref{adding 50}, MSE of the proposed method sharply decreases from around the iteration number 800, and converges to 0 around the iteration number 1,500. On the other hand, the MGU (Const.) converges to 0 around the iteration number 2,700. 
From Fig. \ref{adding 250}, MSE of the proposed method sharply decreases from around 1,700 iterations and converges to 0. In the MGU (Const.), MSE never falls below 0.17.\par

%図\ref{adding 50}と図\ref{adding 250}はadding taskの結果である.
%図\ref{adding 50}より,iteration数800付近から提案手法の平均二乗誤差が急に小さくなり,iteration数1500付近で0に収束する.一方,従来手法は3000手前でやっと0に収束する.
%図\ref{adding 250}より,提案手法の平均二乗誤差がiteration数1700付近から急激に落ち込んで0に収束する.一方で,従来手法は平均二乗誤差の値が0.17より小さくなることはなかった.

Fig. \ref{copy 50} and \ref{copy 200} are the results of the copy task.
The vertical axis represents Softmax Cross Entropy, and the horizontal axis represents the number of iterations.
In both figures, the Softmax Cross Entropy of the MGU (Const.) is smaller than our proposed one up to iteration number 1,000.
After exceeding the number of iterations, the value of the proposed method falls below the value of the MGU (Const.).

%図\ref{copy 50}と図\ref{copy 200}はcopy taskの結果である.
%図\ref{copy 50},図\ref{copy 200}ともにiteration1000までは,従来手法のソフトマックスクロスエントロピーの値は提案手法の値よりも小さい。
%1000iterationを超えた後は提案手法の値が徐々に小さくなっていき,従来手法の値を下回る.

From the above four figures, our proposed method performs better than the MGU (Const.) in these tasks.

%adding task及びcopy taskの図より,Chrono Initializerを導入した提案手法のスコアが,従来手法と比べて良いことが読み取れる。

\section{CONCLUSION}
We adopted Chrono Initializer as the initialization method of MGU in order to improve the performance.
As a result of our experiment, it turned out that our proposed method performed better than MGU (Const.) in the two tasks.
We showed that Chrono Initializer is effective to improve performance of MGU. We would like to conduct experiments that apply the proposed method to machine translation and speech recognition.

% 本研究ではMGUの初期化方法としてChrono Initializerを採用し、性能の向上を目指した。結果として、今回実験を行った2つのタスクにおいて, 提案手法が従来手法の精度を上回ることが確認できた.Chrono InitializerはMGUのパフォーマンス向上に対して有効であることを示した。

\end{document}